\documentclass{article}

\usepackage{graphicx}

\usepackage{latexsym}
\usepackage{listings}

\usepackage{url}

\begin{document}

\title{
Image Processing using miniKanren
}

\author{Niitsuma Hirotaka}



\maketitle

\section{Introduction}

An integral image 
\cite{Viola01robustreal-time}
 is one of the most efficient optimization technique for image processing.
However an integral image is only a special case of a delayed stream\cite{as:sicp} and memoization.
This research discusses generalizing concept of integral image optimization technique, 
and how to generate an integral image optimized program code automatically from abstracted image processing algorithm.

In oder to abstruct algorithms, we forces to miniKanren.
miniKanren\cite{byrd2010relational}
 is a family of programming languages for relational programming.
The book The Reasoned Schemer\cite{TheReasonedSchemer05}
provides a complete implementation in Scheme.
The core of the language fits on two printed pages.
The Scheme implementation of miniKanren is designed to be easily understood.



In this research, we use forked implementation
\footnote{
\url{https://github.com/niitsuma/Racket-miniKanren/tree/recursive}
}
 of miniKanren
from original
miniKanren
\cite{miniKanrenQuine2012}.
implementation
and
Racket language as the scheme language interpreter
\footnote{
\url{http://racket-lang.org/}
}
.

\section{Integral image as 
delayed stream
and
memoization 
}
\label{moving-average}

For the simplicity,
let us consider 
a moving average
\cite{box-blur2001}
on 
one-dimensional data
.
This discussion is generalized to more than two-dimensional case, later.

a
moving average 
 is the unweighted mean of the previous $n$ datum points
as the following.
\begin{lstlisting}[language=Lisp,basicstyle=\footnotesize]
(define (moving-average-simple lst n)
 (let loop1 ((i 0) (result '()))
  (if (>  i (- (length lst) n))
   (reverse result)
   (loop1 (add1 i) 
     (cons
      (/ (let loop2 ((j 0) (sum 0))
	  (if (>= j n) sum
	    (loop2 (add1 j)
	     (+ sum (list-ref lst (+ i j))))))
      n)
     result)))))

(moving-average-simple 
  '(1 2 3 4 5 6) 2)

> '(3/2 5/2 7/2 9/2 11/2)
\end{lstlisting}

A moving average can be implemented using integral image based on delayed stream
as the following
.
\begin{lstlisting}[language=Lisp,basicstyle=\footnotesize]
(define (moving-average-delayed-stream lst n)
  (define lsts (list->stream lst))    
  (define (sum-helper summed lsts)
    (stream-cons 
     (+ (stream-car summed) (stream-car lsts))
     (sum-helper 
      (stream-cdr summed) 
      (stream-cdr lsts))))
  (define summed-table 
    (stream-cons 0 
      (sum-helper summed-table lsts)))
  (define moving-average-stream
    (stream-map
     (lambda (x y) (/ (- x y) n))
     (stream-drop n summed-table)
     summed-table))
  (stream->list  
    (stream-take 
      (- (length lst) n -1)
        moving-average-stream) ))
\end{lstlisting}
However, it is difficult generalizing  this technique to more than tow-dimensional case,
since stream is abstraction of one-dimensional data.

In the case of integral image based on memoization, 
the moving average filter can be implemented
as the following.
\begin{lstlisting}[language=Lisp,basicstyle=\footnotesize]
(require (planet dherman/memoize:3:1))
(define (moving-average-memoize lst n)
 (define/memo (summed-table m)
   (if (<= m 0) 0
     (+ (list-ref  lst(sub1 m))
	(summed-table (sub1 m)))))
 (build-list
  (-(length lst) n  -1)
   (lambda (m)
    (/ (- 
      (summed-table (+ m n))
      (summed-table m)
       ) n))))
\end{lstlisting}
Here
\verb|define/memo| 
defines function 
 \verb|summed-table|
with memoization.
This technique can easily generalize to more than two-dimensional case.

Some time, these
delayed stream and memoization
implementation,
outperforms simple integral image.
For example, let us consider the case only few Haar-like features are sampled.
As is often the case with 
the detected object is near to its position in previous movie frame.
In such case, simple average is faster than integral image,
since summing over whole image contains many redundant computation.
The memoization technique can reduce such redundant computation.

This strategy which selects best optimize,
can be formulated as more general framework using miniKanren.

\section{
Delayed stream 
and
memoization
in miniKanren }

Let us consider the recent imprimentation of miniKanren\cite{miniKanrenQuine2012}.
miniKanren 
can be regarded as 
delayed stream of list of golas
\cite{microKanren2013}.
This raggedness becomes more clear by 
renaming some code blocks as more appropriate name
in 
\verb|lambdaf@|
and 
\verb|case-inf|
macro
as the following.

\begin{lstlisting}[language=Lisp,basicstyle=\footnotesize]
(define-syntax lambdaf@ 
  (syntax-rules () ((_ () e) (delay e))))

(define-syntax case-inf
 (syntax-rules ()
  ((_ e (() e0) ((f^) e1) ((a^) e2) ((a f) e3))
   (let ((a-inf e))
    (cond
     ((not a-inf) e0)
     ((promise?   a-inf) (let ((f^ 
                           (force a-inf))) e1))
     ((not (and (pair? a-inf)
                 (procedure? (cdr a-inf))))
          (let ((a^ a-inf)) e2))
     (else (let ((a (car a-inf)) (f (cdr a-inf))) 
                 e3)))))))
\end{lstlisting}
Note to lines including  
\verb|delay|
,
\verb|force|
and
\verb|promise?|
.
These lines show that 
miniKanren evaluates list of goals as delayed stream.

miniKanren also have auto memoization mechanism.
miniKanren uses triangular substitutions\cite{BaaderSnyder:HandbookAR:unification:2001} .
Triangular substitutions can be regarded 
as generalized memoization mechanism.

Using the following loop statement 
\verb|builde|
in miniKanren
,
these 
delayed stream 
and
memoization  
mechanism
can be included automaticaly.


\begin{lstlisting}[language=Lisp,basicstyle=\footnotesize]
(define (builde n f)
 (let loop ([m 0])
  (if (>= m n) 
    succeed
    (fresh ()
      (f m)
      (loop (add1 m))))))
\end{lstlisting}

\verb|builde|
is the correspondence of 
  \verb|build-list| statement in the scheme language
to miniKanren.
The following example usage shows clearly its usage.

\begin{lstlisting}[language=Lisp,basicstyle=\footnotesize]
(let ([vs (build-list 3 var)])
 (run* (q)
   (builde
     3
     (lambda (i) 	  
      (== (list-ref vs i) i)
    ))
    (== q vs))
)
> '((0 1 2))
\end{lstlisting}

By using 
\verb|builde|
,
moving average filter with 
the strategy which selects best optimize
can be described as the following.

\begin{lstlisting}[language=Lisp,basicstyle=\footnotesize]
(define at list-ref)
(let* (
     [n 5]
     [v (build-list n (lambda(x ) (random n)))]
     [t (build-list n var)]
     [r (build-list n var)]
     [s 2]
      ) 
 (run* (q)		
  (== (at t 0) 0)
  (builde n 
   (lambda (i)
    (fresh (t1) (== t1 (at t i))
     (project (t1)
       (== (at t (add1 i))
	   (+ t1 (at v i)))))))
   (builde
    (- n s -1)
    (lambda (i)
　　　(let ([u (build-list (add1 s) var)])
　　　　(fresh (t1 t2) 
       (== t1 (at t i) )
       (== t2 (at t (+ i s)) )
       (conda

	[
	 (unified-varo t1) (unified-varo t2)
	 (project (t1 t2)
	   (==  (at r i)  (- t2 t1)))
	]

	[
	 (fresh ()
	   (== (at u 0) (at v i))
             (builde
               s
	     (lambda (j)
	       (if (= j (sub1 s))
		  (== (at r i) (at u (sub1 s)))	    
		  (fresh (x)
		    (== x (at u j))
		    (project (x)
		       (== (at u (add1 j))
		       (+ x (at v (+ i j 1)))
　　　　　  )))))))
	]

       )))))	      
       (== q `(,v ,t ,r))
))
\end{lstlisting}

First, 
random signal  
\verb|v|
with length 
\verb|n|
is genereted.
Then 
the signal  
\verb|v|
is averaged over with window length 
\verb|s|.
The avereged result is stored to list 
\verb|r|.
List 
\verb|t|
is integral image of signal 
\verb|v|.
The first loop using 
\verb|builde|
construct integral image.
The moving averages are computed in 
the second 
\verb|builde|
loop.
Most important part is 
inside 
\verb|conda|
.
There are two blocks 
inside 
\verb|conda|
.
The first block
calculates moving average using integral image when 
integral image value is aleady computed.
\verb|unified-varo|
statement judges 
whether or not the value is aleady computed.
When the value not exist,
the second block evaluated.
The second block
simply computes 
moving averages using 
\verb|conda|
loop inside the second block.

Our proposition is the following framework.

\begin{enumerate}
\item describe algorithm  using 
\verb|builde|
loop.

\item when describing algorithm,
enumrate variouse possible relations.

\item sort the variouse relations with sort order its short cut length.

\item enumlate the sorted relation in
\verb|conda|

\item then, 
breast-first search in miniKanren  choses best optimize strategy

\end{enumerate}

This framework provides generalization of integral image optimization strategy.

\section{Nested loop}

\verb|builde|
can define one-dimentinal loop in miniKanren
within 
delayed stream 
and
memoization  
mechanism.
Any dimentional nested loop 
Based on 
\verb|builde|
can define usign the follwing function.

\begin{lstlisting}[language=Lisp,basicstyle=\footnotesize]
(define (builde-nest n-list f)
 (let loop ([n-lst n-list] [i-lst '()] )
   (if (null? n-lst)
    (apply f (reverse i-lst))
    (let ([m (car n-lst)])
     (builde
      m
      (lambda (i)
       (loop (cdr n-lst) (cons i i-lst))))))))
\end{lstlisting}

The following example usage shows clearly its usage.

\begin{lstlisting}[language=Lisp,basicstyle=\footnotesize]
(let ([vs (build-list 6 var)])
 (run* 
  (q)
  (builde-nest
   '(2 3)
   (lambda (i j)
    (== 
     (list-ref vs (+ (* i 3) j ) ) 
     (+ (* i 3) j ))
   ))
   (== q vs)
   ))
 >   '((0 1 2 3 4 5))
\end{lstlisting}

Nothe that, 
\verb|builde-nest|
function 
can convert any nested loop into delayed stream.
There is a possibility that 
appropriate optimization techniques can translate
the converted delayed stream into integral image optimaized code.
However this research forces to abstraction of nested loop.
Developping optimization compiler is future work.


\section{Application to 
connected-component labeling 
}

This section  shows 
example
application of
\verb|builde-nest|
to image processing application.
For the example, we consider 
Connected-component labeling 
\cite{Dillencourt:1992:GAC:128749.128750}
.
Implementing 
Connected-component labeling
in C langage
 need more than 20 lines.
The following is an example code 
taken from
\url{http://stackoverflow.com/questions/14465297/connected-component-labelling}
.

\begin{lstlisting}[language=C,basicstyle=\footnotesize
%,frame=single
]
int dx[] = {+1, 0, -1, 0};
int dy[] = {0, +1, 0, -1};
int row_count;
int col_count;
int m[MAX][MAX];
int label[MAX][MAX];
void dfs(int x, int y, int current_label) {
  if (x < 0 || x == row_count) return;
  if (y < 0 || y == col_count) return;
  if (label[x][y] || !m[x][y]) return;
  label[x][y] = current_label;
  for (int direction = 0; 
       direction < 4; ++direction)
    dfs(x + dx[direction], 
	   y + dy[direction], 
          current_label);
}
void find_components() {
  int component = 0;
  for (int i = 0; i < row_count; ++i) 
    for (int j = 0; j < col_count; ++j) 
      if (!label[i][j] && m[i][j]) 
	dfs(i, j, ++component);
}
\end{lstlisting}

miniKanren with
\verb|builde-nest|
enables 
implementing with one of third lines.

\begin{lstlisting}[language=Lisp,basicstyle=\footnotesize,morekeywords={lambda,run,succeed,img-ref} 
%,frame=single
]
(define result  
 (map var (make-list (* width height) 1)))

(run* (q)
 (builde (list (sub1 height) (sub1 width) 2 2)
  (lambda (i j di dj)
   (if (equal? 
          (img-ref img i j) 
          (img-ref img (+ i di) (+ j dj)))
	(==  
         (img-ref result i j) 
         (img-ref result (+ i di) (+ j dj)))
       succeed)))
 (== q result))
\end{lstlisting}

\begin{figure}
\begin{center}
  \includegraphics[width=2cm]{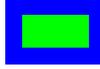}
\end{center}
  \caption{input image}
  \label{fig:rect-img}
\end{figure}

Figure
\ref{fig:connected-component-labeling-miniKanren-result}
shows this programs output when the image in
figure
\ref{fig:rect-img}
is inputed.
This code example shows our framework enables quite readable and 
concise program, especially the algorithms include many nested loop.

\begin{figure}
\begin{lstlisting}[language=Lisp,basicstyle=\footnotesize]
(_.0 _.0 _.0 _.0 _.0 _.0 _.0 _.0 
 _.0 _.1 _.1 _.1 _.1 _.1 _.1 _.0 
 _.0 _.1 _.2 _.2 _.2 _.2 _.1 _.0 
 _.0 _.1 _.2 _.2 _.2 _.2 _.1 _.0 
 _.0 _.1 _.1 _.1 _.1 _.1 _.1 _.0 
 _.0 _.0 _.0 _.0 _.0 _.0 _.0 _.0 )
\end{lstlisting}
\caption{connected-component-labeling result using miniKanren}
\label{fig:connected-component-labeling-miniKanren-result}
\end{figure}



\section{Discussion}

In section \ref{moving-average},
\verb|moving-average-simple|
function is optimized to
\verb|moving-average-delayed-stream|
and
\verb|moving-average-memoize|
using an integral image.
The ultimate aim of this research
is automatic generating these optimized code from simple code.
However,
the proposed framework
is difficult to generete such optimized code.
In order to atuomatic optimize any program, 
the system requires problem speciﬁc knowledge.
It is hard to analyse problem speciﬁc knowledge automatically.
Instead, the proposed framework selects the best relation from already enumerated relations.
When human programmer enumerates relations, problem speciﬁc knowledge is analyized by human programmer.
If this enumerating process can be atomized, we can fully automize this optization process.

Speeded up robust features (SURF)
\cite{Bay:2008:SRF:1370312.1370556}
is optimized algorithm from
scale-invariant feature transform (SIFT)
\cite{Lowe:1999:ORL:850924.851523}
with appropriate approximation.
Approximation is also important factor for integral image optimization.
Also It is hard to analyze approximation automatically.
The proposed framework avoid this automatically analyze by enumerating relations.
The analyze is done when human enumerate relations.


Many algorithms for image processing are based on raster scan over 100 $\times$ 100 pixels.
Since relational programming languages are slow, 
relational programming languages are not suitable for deling with many pixels directory.
Instead, relational programming languages can use for describbing relation among extracted features \cite{batchelor1991intelligent}.
miniKanren is one of the such slow relational programming languages.
miniKanren
is based on delayed stream for list of goals.
Note that we can stop delayed evaluation some intermediate point.
Then we can evaluate the rest evaluation by more optimized Scheme compiler.
In this case, miniKanren works as prepossess optimization.

Recently many Scheme compilers including various optimization techniques have been implemented
\cite{StalinScheme,
Siskind99flow-directedlightweight,GambitSchemeInsideOut}
.
Sometimes, automatic application of these various optimization techniques to C programs 
outperformed hand-written C programs
\cite{Siskind99flow-directedlightweight}
.
These optimized compilers also enables relational programming languages to deling with many pixels directory.

\section{Conclusion}

Automated optimization using an integral image is discussed.
However, this research can not achieve fully automation.
Instead the optimization problem is translated into 
enumerating relations using miniKanren.
The translation based on the alalysis which 
an integral image is only a special case of a delayed stream and memoization.
The proposed framework sometime outperform simple integral image implementation.

\bibliographystyle{miru2014e}
\bibliography{miru2014-eniitsuma}

\end{document}